\pdfoutput=1

\documentclass[11pt]{article}
\usepackage{acl} 

\usepackage{times} 
\usepackage{helvet} 
\usepackage{courier} 
\usepackage{graphicx} 
\usepackage{natbib} 
\usepackage{caption} 
\usepackage{subcaption}
\usepackage{colortbl}

\usepackage{mathtools}
\usepackage{amsmath}
\usepackage{amssymb}
\usepackage{booktabs}
\usepackage[T1]{fontenc}
\usepackage{colortbl}
\usepackage{xcolor}
\usepackage{todonotes}

\usepackage{algorithm}
\usepackage{algorithmic}

\usepackage{newfloat}
\usepackage{listings}

\usepackage{tikz}
\def\checkmark{\tikz\fill[scale=0.4](0,.35) -- (.25,0) -- (1,.7) -- (.25,.15) -- cycle;}

\newcommand\blfootnote[1]{%
  \begingroup
  \renewcommand\thefootnote{}\footnote{#1}%
  \addtocounter{footnote}{-1}%
  \endgroup
}

\title{
Changing Answer Order Can Decrease MMLU Accuracy}

\author{%
Vipul Gupta\textsuperscript{$1,2,\dagger$} \enspace David Pantoja
\textsuperscript{$1,3,\dagger$} \enspace Candace Ross\textsuperscript{$1$} \enspace Adina Williams\textsuperscript{$1$} \enspace Megan Ung\textsuperscript{$1$} \\
{\textsuperscript{$1$} {FAIR at Meta AI}}  \quad\\
{\textsuperscript{$2$} {Pennsylvania State University} }\quad  
{\textsuperscript{$3$} {University of California, Berkeley
} }\quad  \\
\quad\\
\normalsize{\tt vkg5164@psu.edu \quad davidpantoja@berkeley.edu}
\\ \normalsize{\tt \{ccross, adinawilliams, meganu\}@meta.com}\\
}

\begin{document}
\maketitle

\begin{abstract}
As large language models (LLMs) have grown in prevalence, particular benchmarks have become essential for the evaluation of these models and for understanding model capabilities. Most commonly, we use test accuracy averaged across multiple subtasks in order to rank models on leaderboards, to determine which model is best for our purposes. In this paper, we investigate the robustness of the accuracy measurement on a widely used multiple choice question answering dataset, MMLU. When shuffling the answer label contents, we find that all explored models decrease in accuracy on MMLU, but not every model is equally sensitive. These findings suggest a possible adjustment to the standard practice of leaderboard testing, where we additionally consider the percentage of examples each model answers correctly by random chance.
\end{abstract}

\blfootnote{$\dagger$ This work was done during their internships at FAIR, Meta.}

\section{Introduction}\label{sec:intro}


One of the largest outstanding issues with interpreting the results of model evaluation pertains to the robustness of accuracy measurements. 
For example, NLP model accuracy has been shown to be fairly brittle. For example, accuracy can drop when researchers apply input alterations based on paraphrasing \citep{gan-ng-2019-improving}, word order changes \citep{gauthier-levy-2019-linking, ribeiro-etal-2020-beyond, sinha-etal-2021-masked, sinha-etal-2022-curious, allenzhu-li-2023-physics31, allenzhu-li-2023-physics32, berglund-etal-2023-reversal, golovneva-etal-2024-reverse, kitouni-etal-2024-factorization, sugawara2020assessing}, or other minor, largely meaning-preserving input variations or perturbations \citep{belinkov-bisk-2018-synthethic, ebrahimi-etal-2018-hotflip, jiang-etal-2020-know, gao-etal-2021-making, li-etal-2021-contextualized, sinha-etal-2021-perturbing, moradi-samwald-2021-evaluating, papakipos-bitton-2022-augly, qian2022perturbation, goodarzi-etal-2023-robustness, sinha-etal-2023-language}. If many models fail to be robust on a benchmark, regardless of their initially measured accuracy, we may need to reconsider how we use it as the basis for a leaderboard that actually ranks models. 

While there are many approaches to investigating robustness, our approach relies on the intuition that a test-taker, human or model, should always select the right answer regardless of its label, i.e. whether it is listed as answer `A' or `C'. Surely, if the right answer is unknown to the test-taker and they make an uneducated guess, they still could happen upon the right answer by chance, but, in an ideal scenario, a true expert should achieve the same score when tested multiple times on versions of a test where only the order that answers are presented in changes.  

In humans, this performance stability, often called \textbf{test-retest reliability} is an important consideration to determine how to interpret the results of running a test \citep{bland-altman-1986-statistical}. Humans test scores can fluctuate over time, because they are filtered through irrelevant mental or physical factors that affect measurement \citep{spearman-1910-correlation, dunlap-1933-comparable}. Such uninformative fluctuations can affect multiple choice tests, for example, when answers are presented in a different order during retest \citep{krosnick-fabrigar-1991-handbook, tellinghuisen-2008-does, lions-etal-2022-does}. However, as models do not have the biological limitations of humans, we may expect them to exhibit less variation than humans, or possibly even none at all. Thus, we claim that a model should be robust to answer order changes: if it gets the correct answer to a question when the answer is labeled `A', it should also always get the correct answer when it is labeled `C'. Put another way, the model should select the same answer for each question, regardless of its label, for every possible version of a benchmark; its accuracy should be static between test and retest.

In our work, we ask whether shuffling the order of the answer label contents, leaving the order of the labels (A, B, C, D) the same, affects the measurement of accuracy. We focus our investigation on the MMLU dataset, a popular dataset included on the widely used Hugging Face Open LLM Leaderboard\footnote{\url{https://huggingface.co/spaces/open-llm-leaderboard/open_llm_leaderboard}}, which runs with the Eleuther LM Evaluation Harness \citep{gao-etal-2023-evalharness} as its backend. 

Testing top performers on the Open LLM Leaderboard, we find that all ten models are affected by our answer shuffling. This indicates that serious non-robustness in benchmarking with MMLU. To better rank models on a leaderboard with the MMLU dataset, we may want to take more random shuffles of label contents to better understand the extent to which a model genuinely can output the correct answer.   

\section{Methods}\label{sec:methods}

\subsection{MMLU}\label{subsec:data}
Massive Multitask Language Understanding (MMLU) is a commonly used benchmark for evaluating LLMs \citep{hendrycks2021measuring}. It is intended to test a model's world knowledge and problem solving ability, and consists of 57 tasks. Each example in MMLU consists of a question paired with four possible answers, only one of which is correct. Answers are a concatenation of an answer label denoted as a letter, with answer contents (a string of characters). To test the robustness of models to answer choice ordering, we shuffle the answer label contents, with prohibition that the correct answer contents don't change and that we preserve the ordering of MMLU answer labels (A, B, C, D) across different evaluation runs, for example: 

\begin{center}
\begin{tabular}{ c c c c c c }
 \multicolumn{3}{c}{original} & \multicolumn{3}{c}{a possible shuffle}  \\ 
 &  A. 1 &            & A. 4 & &\\  
 &  B. 2 &            & B. 2 & &\\  
 &  C. 3 & \checkmark & C. 1 & &\\  
 &  D. 4 &            & D. 3 & \checkmark &\\  
\end{tabular}
\end{center}

We can think of the original orders of answer content labels in each example in MMLU as one of the $n$ (out of 24 possible) shuffles of the example. Given the size of the MMLU dataset, it is not efficient to run all the possible shuffles (as each example has 24 options and there are nearly 14 thousand questions. To do a tractable exploration, we take two random seeds of MMLU, each of which has been shuffled, where each example has been selected from one of the 24 possible answer contents orders to create semantically equivalent versions of MMLU. We utilize the original MMLU implementation \citep{hendrycks2021measuring}, which uses 5-shot in context learning during evaluation. 
 
 

\begin{figure*}[ht]
\begin{center}
\includegraphics[width=0.9\linewidth]{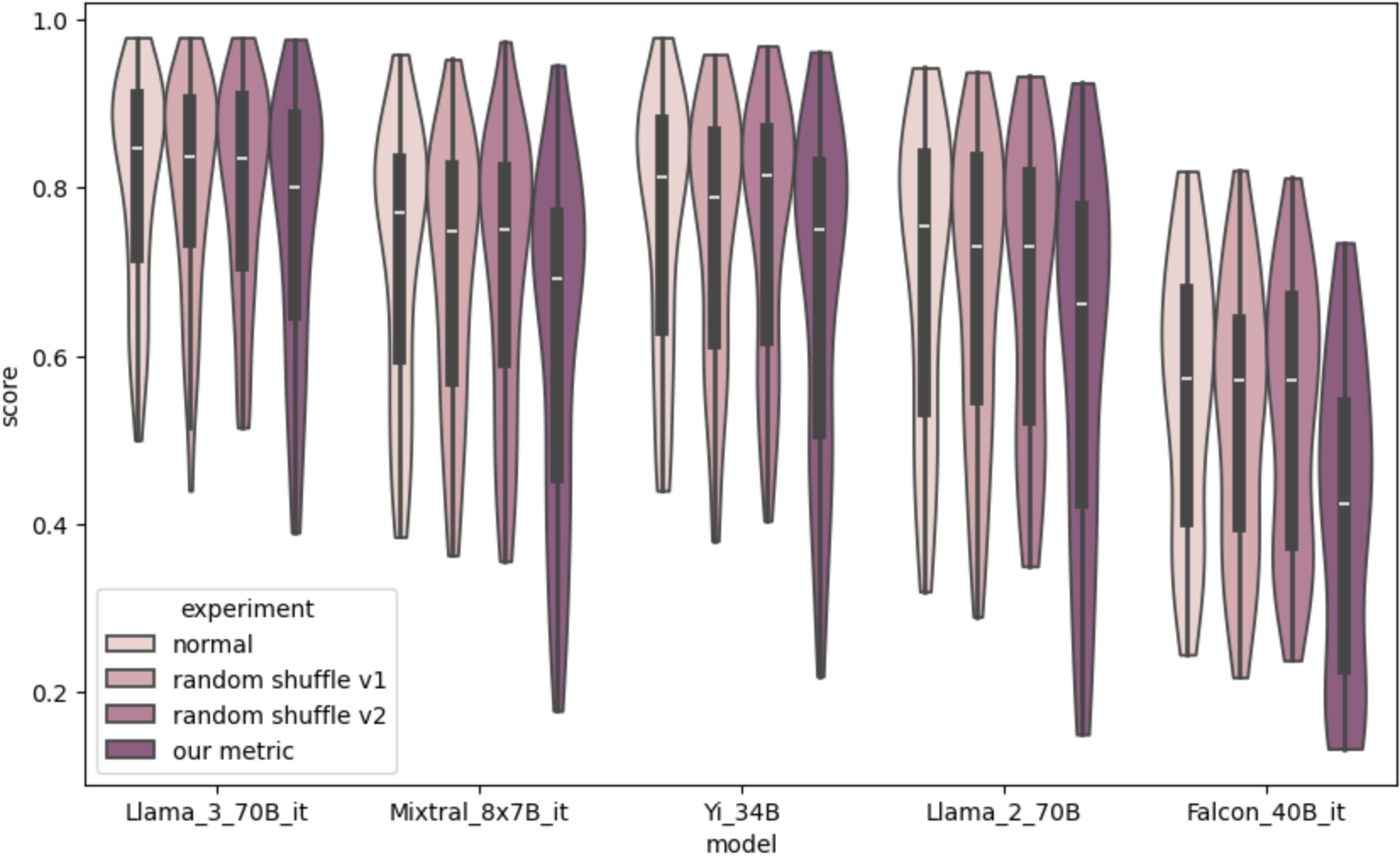}
\end{center}
  \caption{
  This figure illustrates the performance of a selection of state-of-the-art models that we tested on the original MMLU (v0) and 2 shuffled versions (v1 and v2). Models are ordered by accuracy drop in `our metric'. Here `-it' denotes an instruction tuned model. The width of the violin corresponds to the number of subdatasets where the model received a particular score. The white indicator marks the median score for subdataset accuracies. }
\vspace{-1em}
\label{fig:main}
\end{figure*}

\subsection{Metrics}

In essence, we adopt a simplification of the classic formulation of test-retest repeatability from \citeauthor{bland-altman-1986-statistical} to match the ML leaderboard setting: an evaluation (the running of a test on a model) is deemed perfectly stable, if and only if the measurements realized at one time of running it produces \emph{the same exact values} when repeated at a later time, when the test is run under the same conditions. We minimally alter the testing conditions when we repeat the test to measure robustness---by changing the order of answer contents---but all other testing parameters remain static. In our setting, we set the number of test takers, $n$, to 1. 

In simple terms, this metric measures how often the model answers the questions correctly in both the original and the shuffled versions. If the model is actually robust, it will select the right answer no matter where it appears, as the answer's meaning doesn't change when you merely change its label and location in the answer string. If the model's accuracy does change in this setting, then we can say the model isn't actually very competent on the task that the test is testing. 

To quantify (non-)robustness to answer order shuffling, we define a new metric, \textit{our metric}, which measures how often the model answers the same question(s) correctly in both the original and in a shuffled version of MMLU. We take the average over all the shuffles performed as our metric: 

\begin{equation}
\text{Our Metric} = \frac{1}{N}\sum_{i=1}^{N}
\frac{1}{M} \sum_{j=1}^{M} V_{0}^{i}V_{j}^{i} ,
\end{equation}

where $V_{0}^{i} \in \{0,1\}$ indicates whether the model answers the $i^{th}$ question correctly in MMLU dataset (1 if correct, 0 if incorrect). $V_{j}^{i}$ indicates whether the model answers  $i^{th}$ question correctly in the $j^{th}$ shuffled version of the answer label content. $M$ is the total number of shuffles in the scope of the experiment (for us 2) and $N$ is the dataset size. We then take the average performance across two such shuffles.

As formulated, our metric tries to capture the true capabilities of the model by reducing the number of questions correctly answered by random chance. Assuming models do not have external memory of earlier queries, enforcing that the model correctly identify the answer $M$ times (for us twice), noticeably lowers the chance of it happening across the correct answer by chance.

\subsection{Models}
In this work, we evaluate 10 state-of-the-art LLMs, ranging in size from 7 billion to 70 billion parameters, most of which have performed very well on the Hugging Face Open LLM leaderboard. The 10 models we use are: Llama3 70B Instruct, Llama3 70B,  Llama3 8B Instruct \citep{meta-Llama3}, Llama2 70B \citep{touvron2023Llama},  Yi 34B \citep{ai2024yi}, Mixtral 8x7B and Mixtral 8x7B Instruct \citep{jiang2024mixtral}, Falcon 40B Instruct \citep{almazrouei2023falcon}, Mistral 7B Instruct \citep{jiang2023mistral}, and Gemma 7B Instruct \citep{gemmateam2024gemma}. All models are openly available, which enables the reproducibility of our findings.

\begin{figure*}[t]
  \begin{subfigure}[ht]{0.49\textwidth}
    \includegraphics[width=\textwidth, height=5cm]{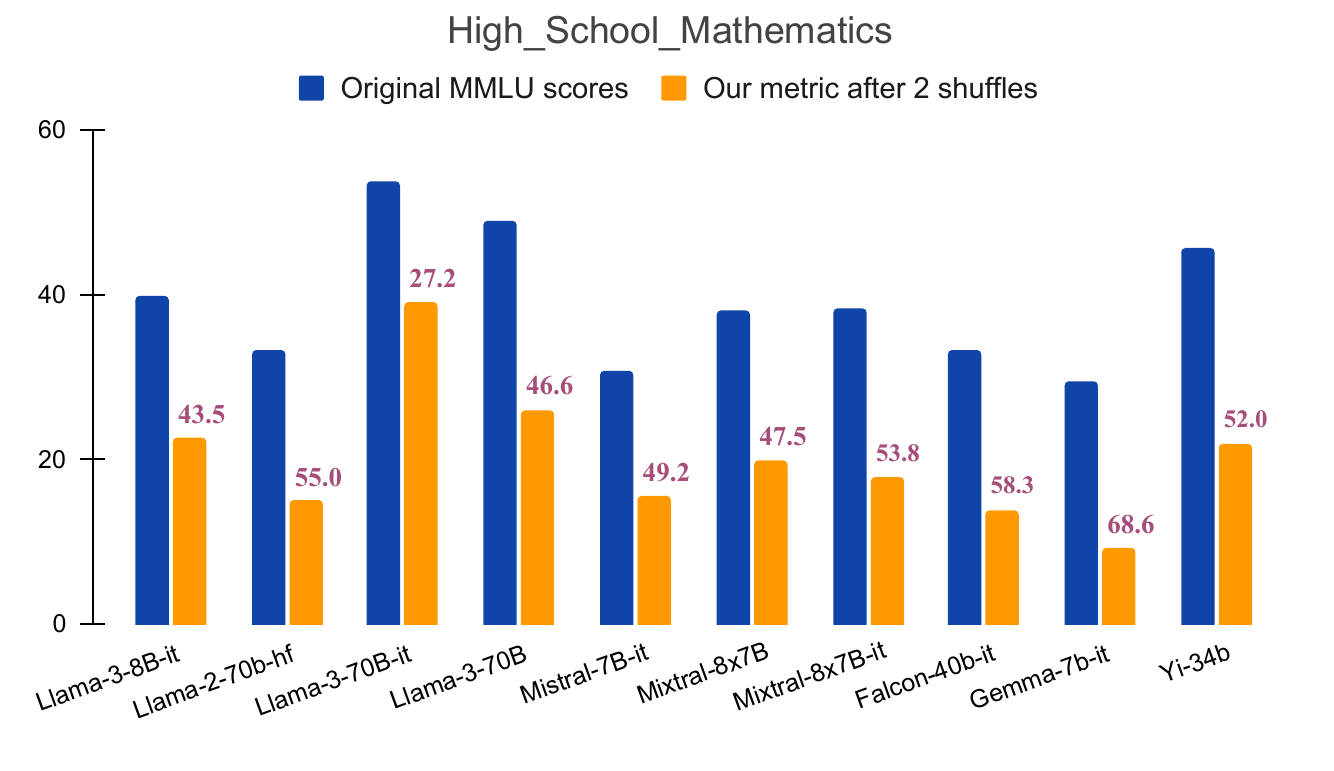}
    \caption{Category of MMLU most affected}
    \label{fig:figure1}
  \end{subfigure}
  \hspace{0.02\textwidth}
  \begin{subfigure}[ht]{0.49\textwidth}
    \includegraphics[width=\textwidth, height=5cm]{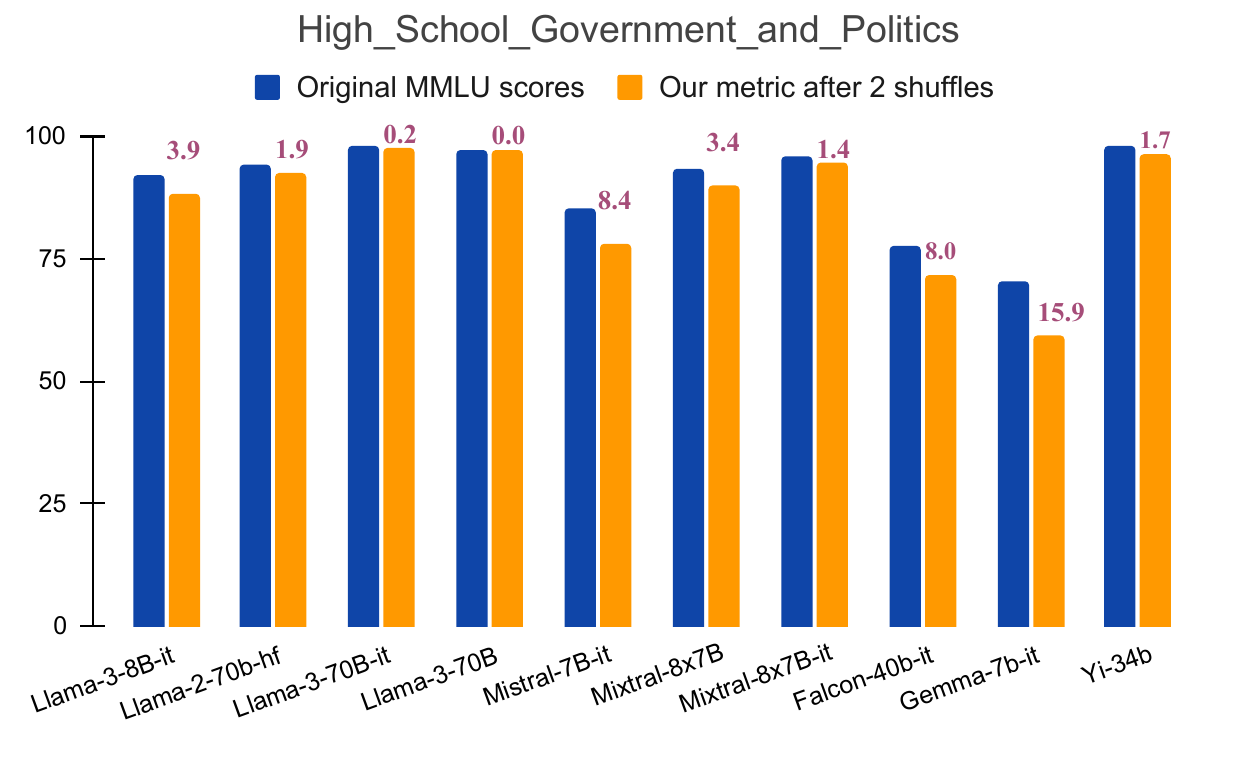}
    \caption{Category of MMLU least affected}
    \label{fig:figure2}
  \end{subfigure}
  \caption{The most and least affected categories of MMLU with our proposed shuffling. The number above each plot signifies percentage change after shuffling. Here `-it' marks instruction finetuned models.}
  \label{fig:category_plots}
\end{figure*}

\section{Results}\label{sec:results}

We found that all tested models performed worse according to our metric after answer content shuffling than on the original version of the dataset, as shown in Table \ref{tab:percent_drop}. After shuffling, we see that models fail to select the correct answer for every question it originally selected correctly, as shown by \textit{our metric} in Figure \ref{fig:main}. 

\begin{table}[ht]
\small
\centering
\begin{tabular}{lrrr}
\toprule
\textbf{Model Name} & {\hspace{-0.5em}}\textbf{MMLU}{\hspace{-0.5em}} & {\hspace{-0.5em}}\textbf{Our Metric} & {\hspace{-0.5em}}\textbf{\% Drop} \\ \midrule
Llama-3-70B-it &  80.3    &   75.3    &   \cellcolor[RGB]{255,220,0}6.2  \\ 
Llama-3-70B    &  78.9    &   72.4    &   \cellcolor[RGB]{255,200,0}8.2  \\ 
Yi-34B         &  75.8    &   67.7    &   \cellcolor[RGB]{255,180,0}10.7 \\  
Mixtral-8x7B-it&  70.6    &   60.7    &   \cellcolor[RGB]{255,150,0}14.0 \\ 
Mixtral-8x7B   &  70.4    &   60.9    &   \cellcolor[RGB]{255,160,0}13.5 \\ 
Llama-2-70B    &  69.0    &   58.8    &   \cellcolor[RGB]{255,140,0}14.8 \\ 
Llama-3-8B-it  &  66.4    &   58.0    &   \cellcolor[RGB]{255,170,0}12.7 \\ 
Mistral-7B-it  &  59.3    &   46.5    &   \cellcolor[RGB]{255,140,140}21.6 \\ 
Falcon-40B-it  &  54.7    &   39.8    &   \cellcolor[RGB]{255,100,100}27.2 \\ 
Gemma-7B-it    &  51.7    &   38.0    &   \cellcolor[RGB]{255,120,120}26.5 \\ \bottomrule 
\end{tabular}
\caption{Accuracy drop on MMLU due to changing answer order. Here `-it' marks instruction tuned models.}
\label{tab:percent_drop}
\end{table}

We find that some models had higher retest accuracy than others. Models from the Llama-3 family were the most robust, especially Llama-3-70B. Interestingly, Llama-3-8B model was more robust than larger, generally high performing models such as Mixtral-8x7B and Llama-2-70B. For Llama3-70B and Mixtral-8x7B, we also found that their base and instruction finetuned models were comparably robust. Smaller models, like Mistral-7B and Gemma-7B, were generally more impacted. This result is consistent with findings in \cite{zhou-etal-2024-revisiting-self}, which found more inconsistency for smaller models (less than 8B parameters), although in a slightly different setting.
Some larger models, such as Falcon-40B-instruct whose score dropped from 54.7 to 39.8 with our approach, were also strongly impacted. 

We also analyzed performance drop by subdataset in Table~\ref{tab:problem_solving}, and discovered that the models struggled the most with problem-solving subdatasets, such as high school mathematics. For Gemma-7B and Falcon-40B models, the drop in accuracy on these categories were as high as 40\%. As these subdatasets make up a significant portion (over 15\%) of original MMLU dataset, this analysis suggests serious robustness issues affecting accuracy scores on problem-solving categories. Additionally, among most impacted subdatasets, such as ``college mathematics'' and ``global facts'', we investigated whether the drop may be due to the fact that shuffling can ablate the logical order of the original questions. In humans, presenting answer orders in logical order---such as 0,1,2,3 or 3,2,1,0---is recommended by test design research, because random order may pose unnecessary challenge for lower ability students \citep{huntley-welch-1993-numerical, haladyna-etal-2002-review}. We discovered that more than 95\% of the original MMLU dataset was presented in logical order, which indicates that models may be benefiting from logical answer order and perhaps that they should be seen as lower ability test takers.

\begin{table}[ht]
\small
\centering
\begin{tabular}{lrrr}
\toprule
\textbf{Model Name} & {\hspace{-0.5em}}\textbf{MMLU}{\hspace{-0.5em}} & {\hspace{-0.5em}}\textbf{Our Metric} & {\hspace{-0.5em}}\textbf{\% Drop} \\ \midrule
Llama-3-70B-it &  72.1    &   64.5    &   \cellcolor[RGB]{255,180,0}10.5  \\ 
Llama-3-70B    &  68.7    &   57.7    &   \cellcolor[RGB]{255,140,0}16.0  \\ 
Yi-34B         &  65.6    &   52.9    &   \cellcolor[RGB]{255,120,0}19.4 \\ 
Mixtral-8x7B-it&  56.9    &   43.4    &   \cellcolor[RGB]{255,140,140}23.7 \\ 
Mixtral-8x7B   &  57.0    &   43.4    &   \cellcolor[RGB]{255,120,120}23.9 \\ 
Llama-2-70B    &  54.6    &   40.4    &   \cellcolor[RGB]{255,100,100}26.0 \\ 
Llama-3-8B-it  &  54.3    &   40.9    &   \cellcolor[RGB]{255,110,110}24.7 \\ 
Mistral-7B-it  &  45.2    &   29.8    &   \cellcolor[RGB]{255,70,70}34.1 \\ 
Falcon-40B-it  &  41.5    &   24.3    &   \cellcolor[RGB]{255,60,60}41.4 \\ 
Gemma-7B-it    &  38.9    &   22.2    &   \cellcolor[RGB]{255,50,50}42.9 \\ \bottomrule
\end{tabular}
\caption{Accuracy drop on problem solving categories of MMLU dataset due to option text shuffling. 
}
\label{tab:problem_solving}
\end{table}

\section{Related Works}

\paragraph{LLMs can be Sensitive to Option Order and Label.}
Recent works have also shown that the accuracy of models on multiple-choice question datasets can change significantly when the order of answer options is rearranged \cite{robinson2023leveraging, pezeshkpour-hruschka-2024-large, alzahrani-etal-2024-benchmarks, wei-etal-2024-unveiling, xue2024strengthened, zong2024fool}. This suggests that models are sensitive to the order of answer options, which can impact their performance. \cite{wang2024beyond} studies how changing the number of options in mutliple choice question datasets affect the model performance. They found that LLMs have overfitted to multiple choice question datasets with exactly four options. 

Other studies have shown that models may exhibit prior biases towards specific option IDs (e.g., `A') \cite{wei-etal-2024-unveiling, zheng2023judging, reif-schwartz-2024-beyond, ross-etal-2024-what, li2024anchored, zheng2023large}. Some works have also shown that models can perform surprisingly well above random chance even when question text is removed and only answer options are provided \cite{balepur-etal-2024-artifacts, shah-etal-2020-expect, balepur-rudinger-2024-large}. Recent works have also shown that replacing the correct option with ``None of the Above'' leads to a drastic decline in performance across all models \citep{wang2024beyond, xu2024llms}. These findings suggest that models may be relying on artifacts or biases in the data rather than truly understanding the questions \cite{rottger2024political, raj2023semantic}.

In a concurrent work, \citet{mcilroyyoung2024setbasedpromptingprovablysolving} proposed a solution fro mitigating the issue of order dependency in LLMs by modifying the self attention matrix of the input sequence. They set the attention scores between different options to be zero, effectively preventing the model from attending to the order of options. 

In contrast to above works, our work focuses on category-wise differences in model performance and proposes a new metric that takes into account the variation in model performance across different answer orders. Our approach provides a more nuanced understanding of the impact of answer option ordering on model accuracy.

\paragraph{Evaluation Dataset Validity.}
For all evaluation datasets, validity is important, and MMLU is no exception. Several recent works have discussed MMLU's validity \citep{gema-etal-2024-we, zheng2023large, wang2024beyond, wang2024mmlu}. In particular, \citet{wang2024mmlu} found trivial and noisy questions in the dataset and proposed an update, MMLU-Pro, which aims to mitigate those issues. 
Concurrent work on model robustness to question-answering order \citep{zhou-etal-2024-revisiting-self} applies a similar approach to ours that shuffles answer label content and also explores other possible modes of interrogating robustness. While they also find non-robustness to question variants, our work differs from theirs in that our metric can account for the multiplicity of potential orderings of answer labels; we provide further analysis for each category in MMLU in \autoref{fig:appenedix_plots} in the appendix.

\section{Conclusion}\label{sec:discconclusion}


This work tested the robustness of the evaluation benchmark pipeline for the popular leaderboard dataset MMLU. To separate out the effect of chance on model answers, we apply a largely meaningless change to the datasets by shuffling label contents. We find that this meaning-preserving alteration resulted in a decrease in MMLU accuracy for all models, but not to the same degree. We define a new metric that quantifies the effect of chance and suggest that it is important to take it into consideration during evaluation and leaderboard rankings of models.




\bibliography{references, anthology}
\bibliographystyle{acl_natbib}

\appendix

\section{Appendix}
\label{sec:appendix}

\subsection{Limitations} While we explore two possible shuffles of the answer label contents, we restricted ourselves to the $M$ to curtail compute costs. We do acknowledge that there are many more possible shuffles that might be tested, and more would doubtless lead to a better approximation of the non-robustness. 

\subsection{Category Wise Analysis}
We analyzed how changing the answer order affects each category in the MMLU dataset. We found that some categories are more sensitive to these changes than others. \autoref{fig:category_plots} shows the impact of answer order changes on eight randomly selected categories.

The MMLU has 57 subcategories, and we observed that some categories are more affected by answer order changes than others. For example, categories such as high school physics, abstract algebra, college mathematics, and moral disputes witnessed a significant decrease in performance after answer order changes. On the other hand, categories such as high school us history, econometrics, and professional law were less affected. In some cases, the impact was highly significant - for instance, the accuracy for Mistral-7B-instruct model on moral scenarios category decreased by 77\%, from 31.4 to 7.1, after changing the answer order.

The different plots in \autoref{fig:category_plots} highlight that not all categories are equally affected, some parts of MMLU dataset might be good indicator of model performance.

\begin{figure*}[t]
  \begin{subfigure}[ht]{0.49\textwidth}
    \includegraphics[width=\textwidth, height=5cm]{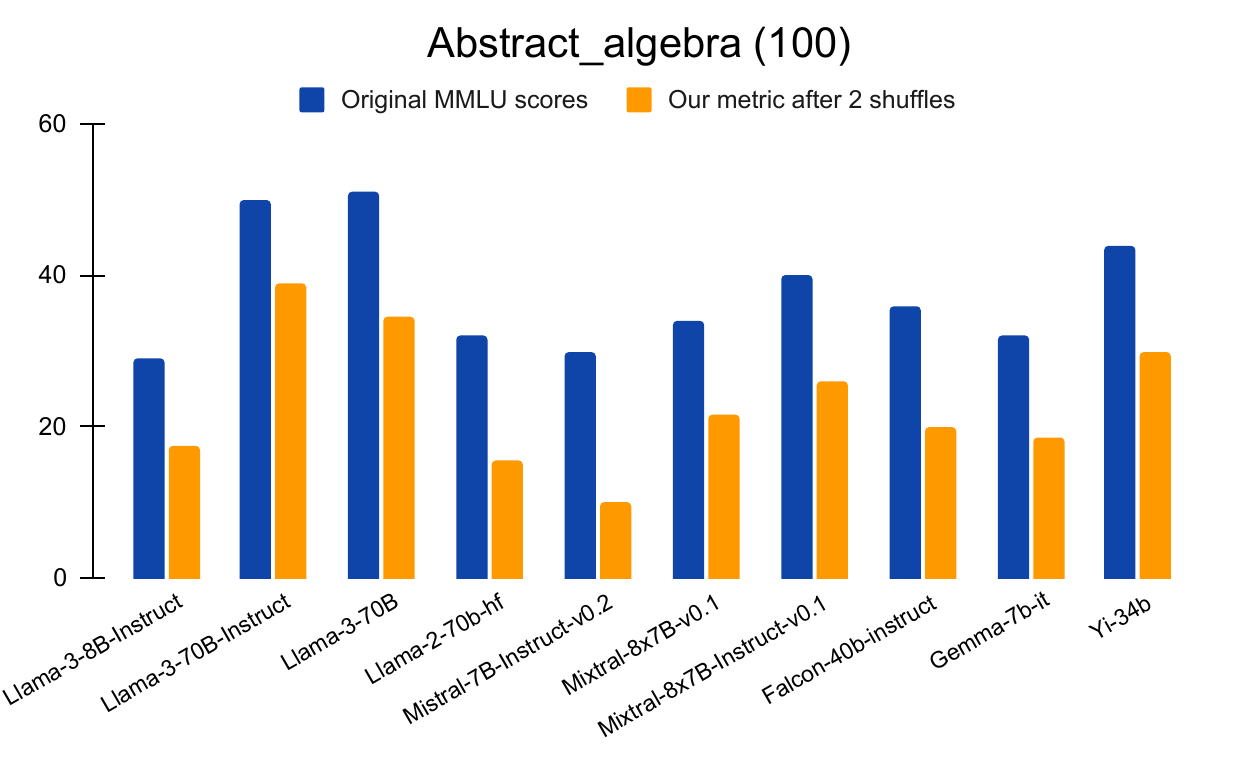}
  \end{subfigure}
  \begin{subfigure}[ht]{0.49\textwidth}
    \includegraphics[width=\textwidth, height=5cm]{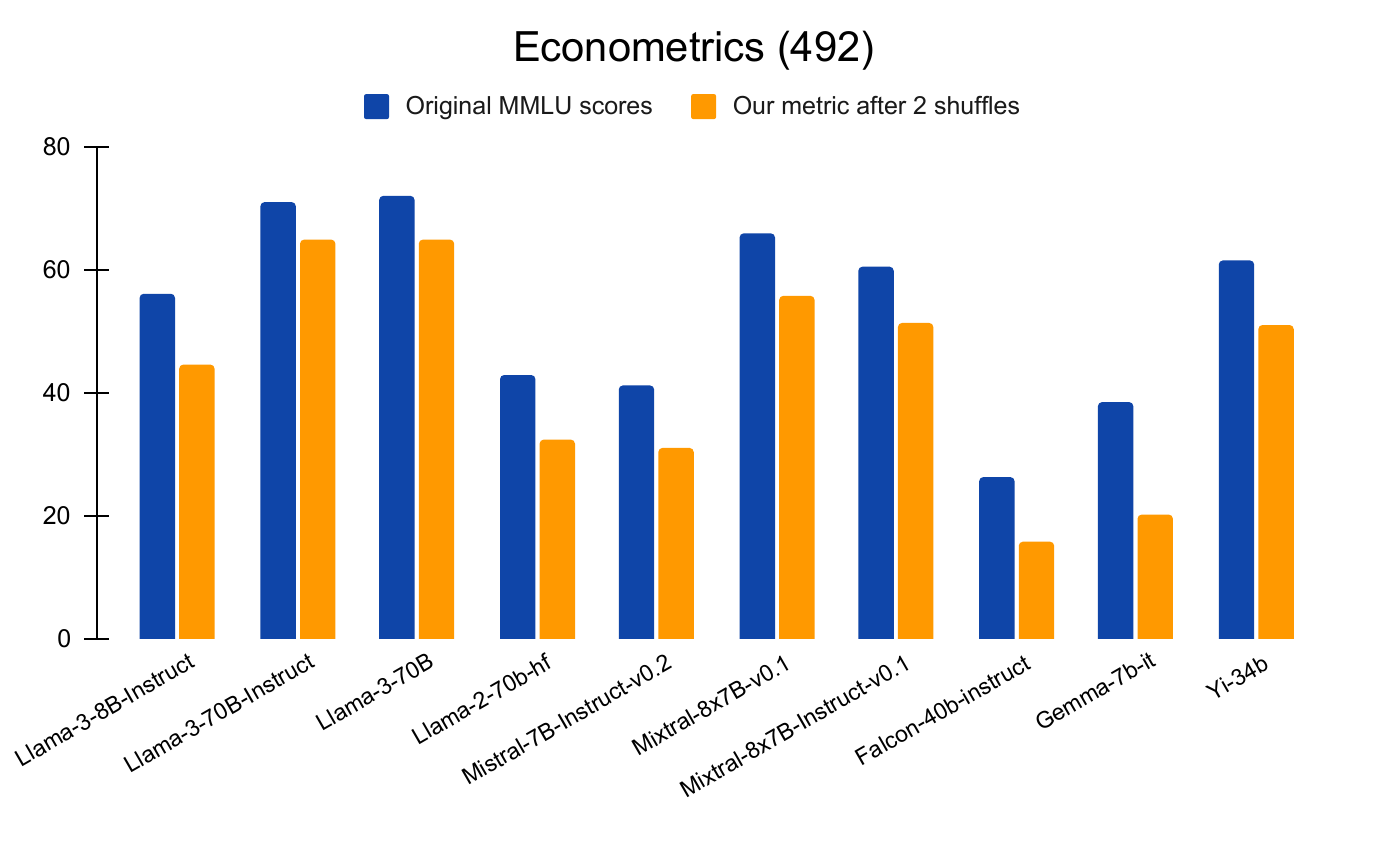}
  \end{subfigure}
    \begin{subfigure}[ht]{0.49\textwidth}
    \includegraphics[width=\textwidth, height=5cm]{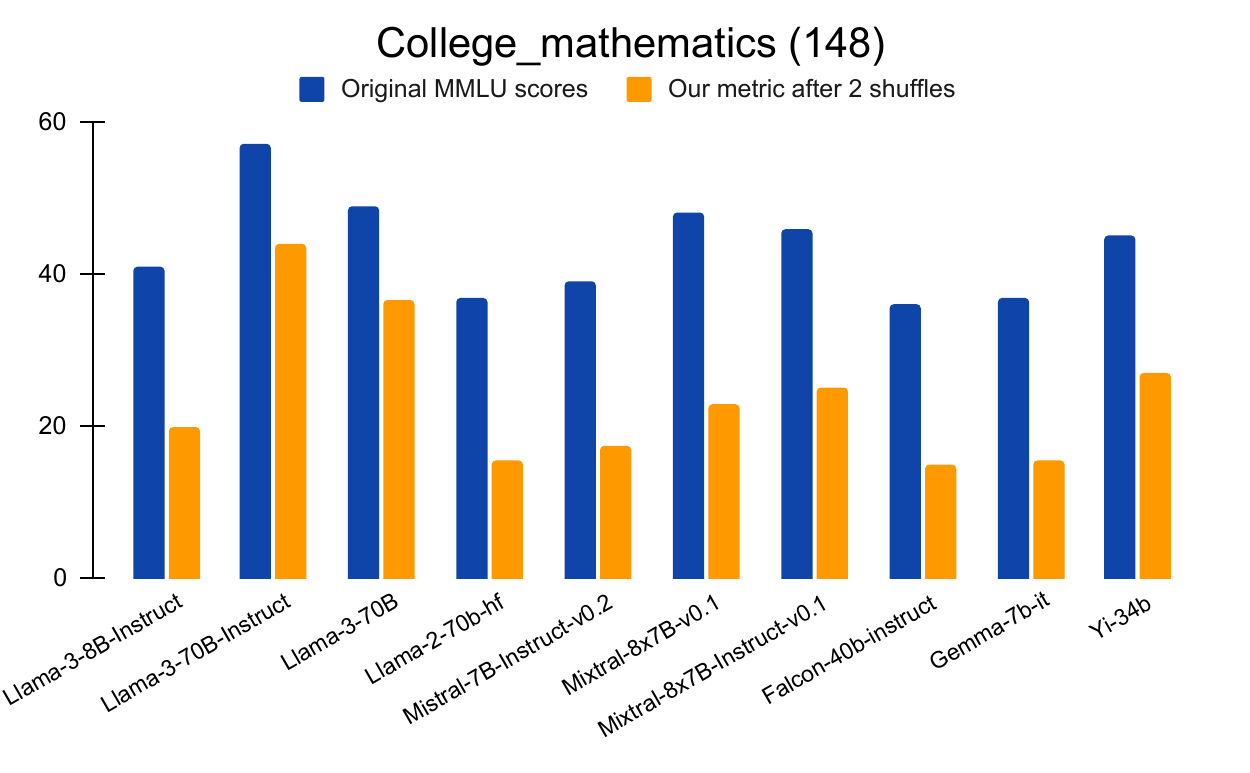}
  \end{subfigure}
  \begin{subfigure}[ht]{0.49\textwidth}
    \includegraphics[width=\textwidth, height=5cm]{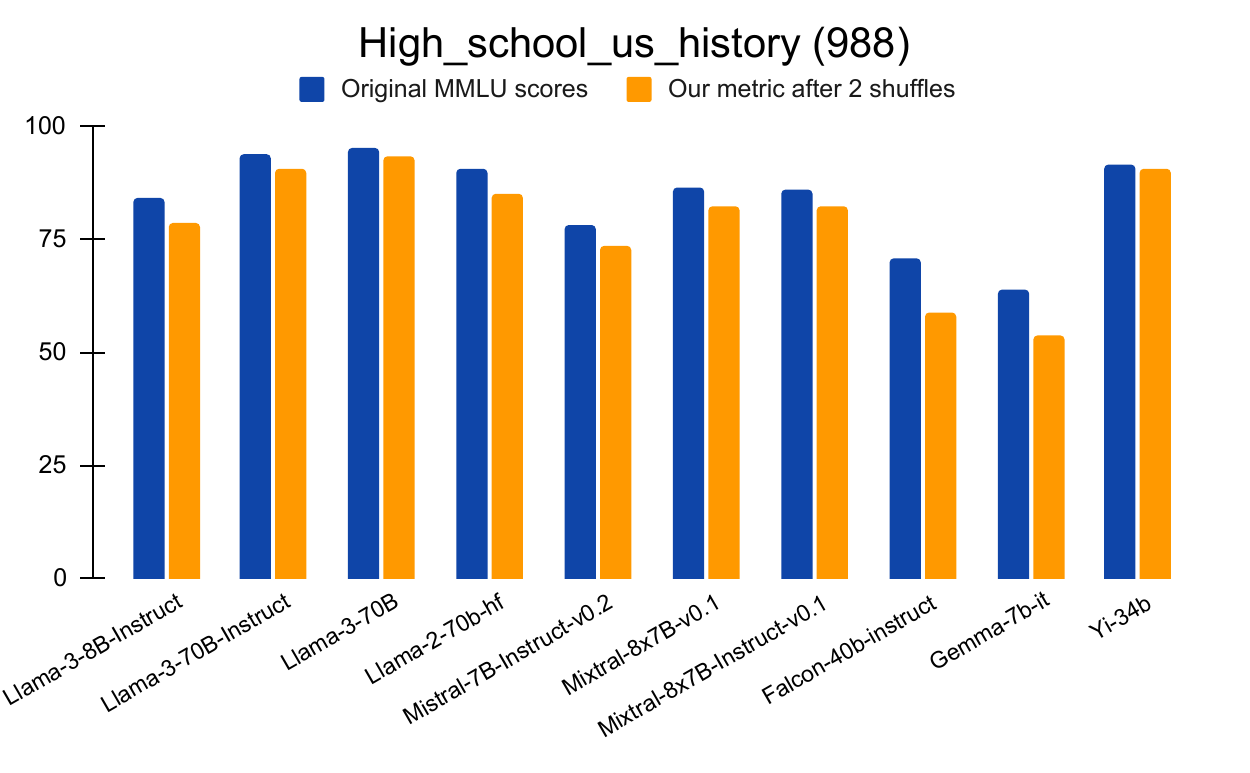}
  \end{subfigure}
  \begin{subfigure}[ht]{0.49\textwidth}
    \includegraphics[width=\textwidth, height=5cm]{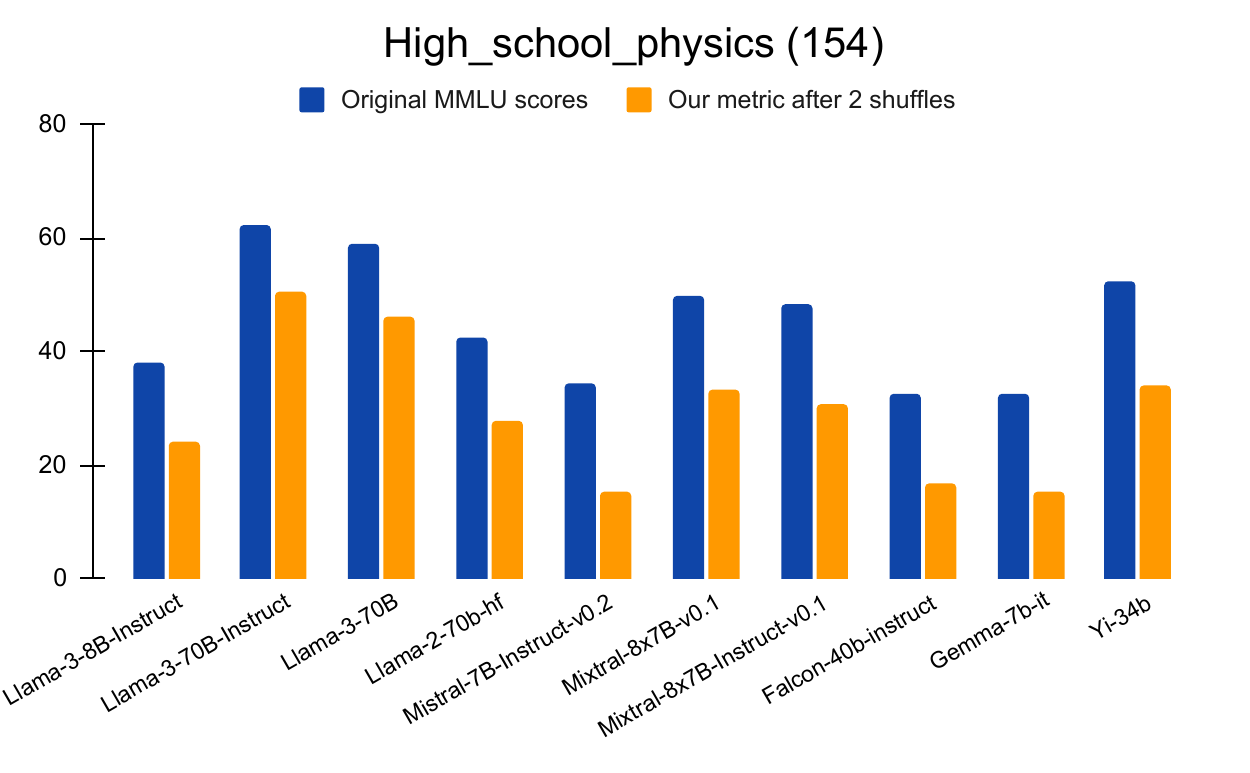}
  \end{subfigure}
  \begin{subfigure}[ht]{0.49\textwidth}
    \includegraphics[width=\textwidth, height=5cm]{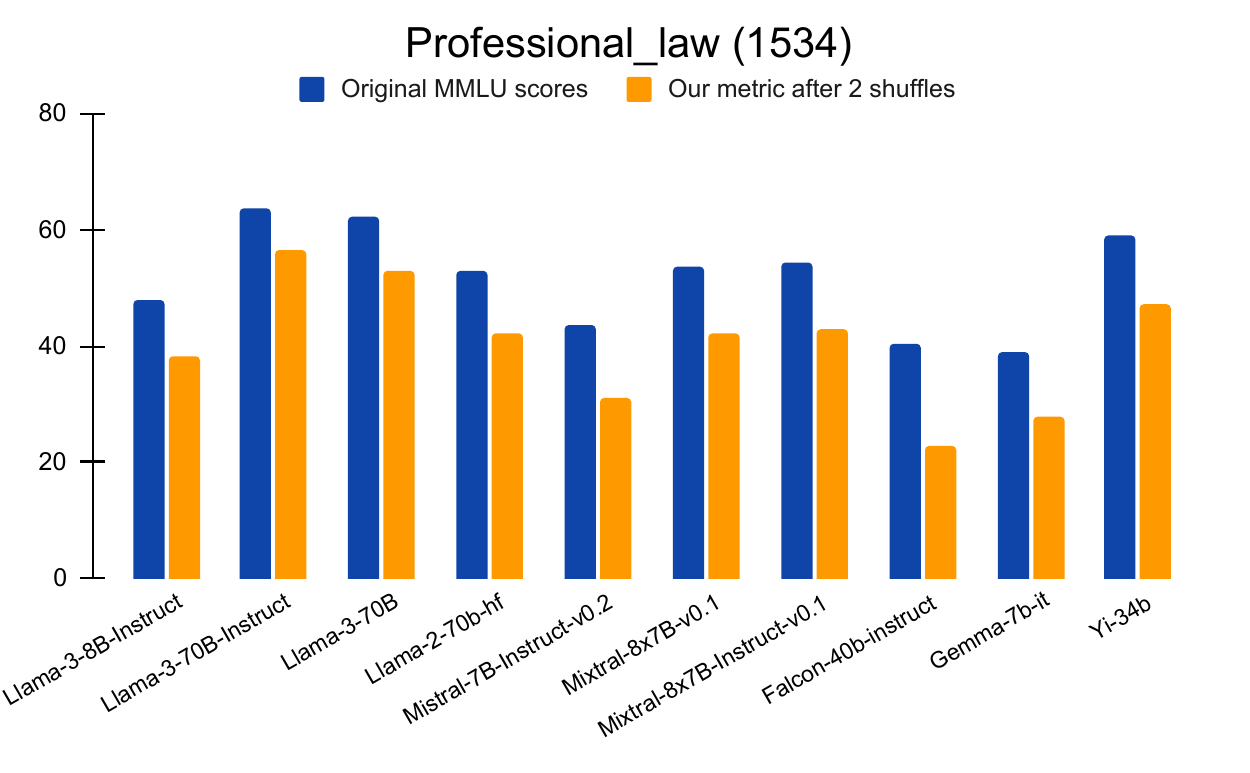}
  \end{subfigure}
    \begin{subfigure}[ht]{0.49\textwidth}
    \includegraphics[width=\textwidth, height=5cm]{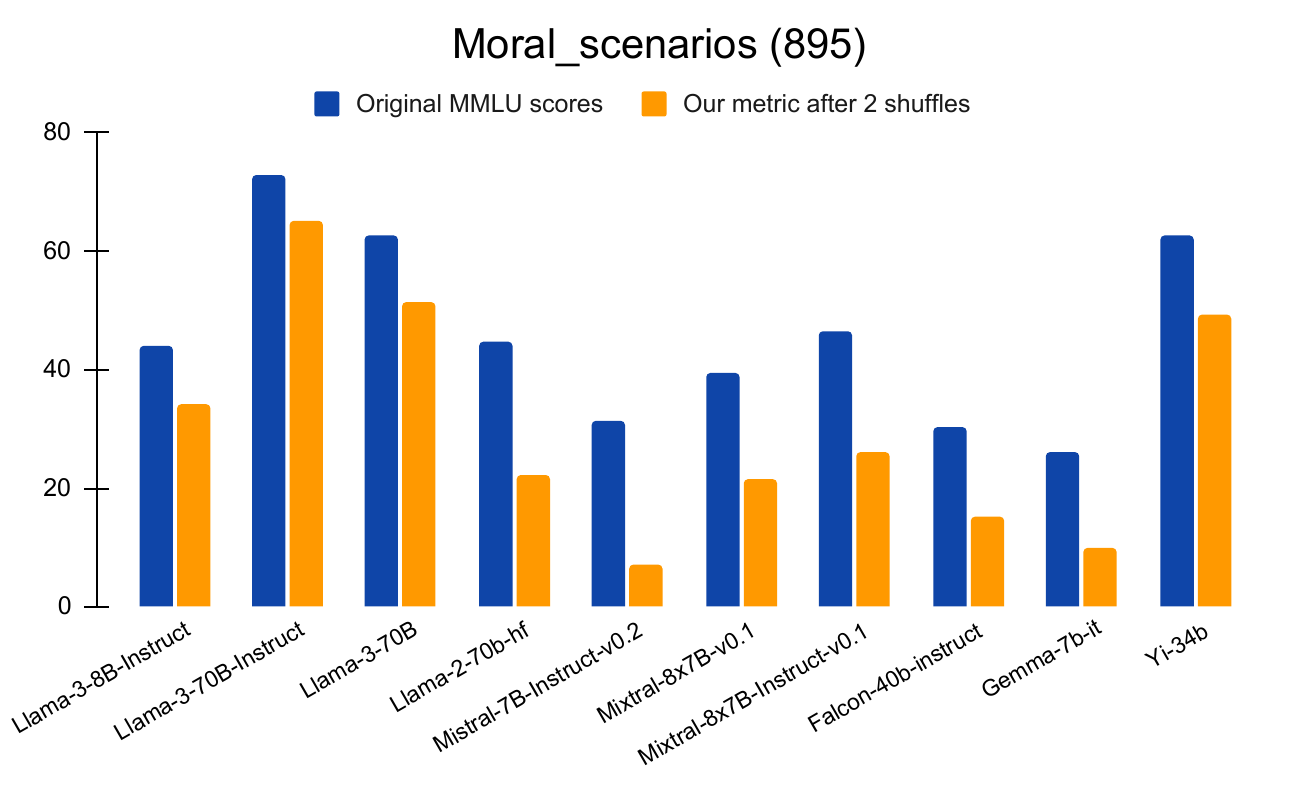}
  \end{subfigure}
  \hspace{0.02\textwidth}
  \begin{subfigure}[ht]{0.49\textwidth}
    \includegraphics[width=\textwidth, height=5cm]{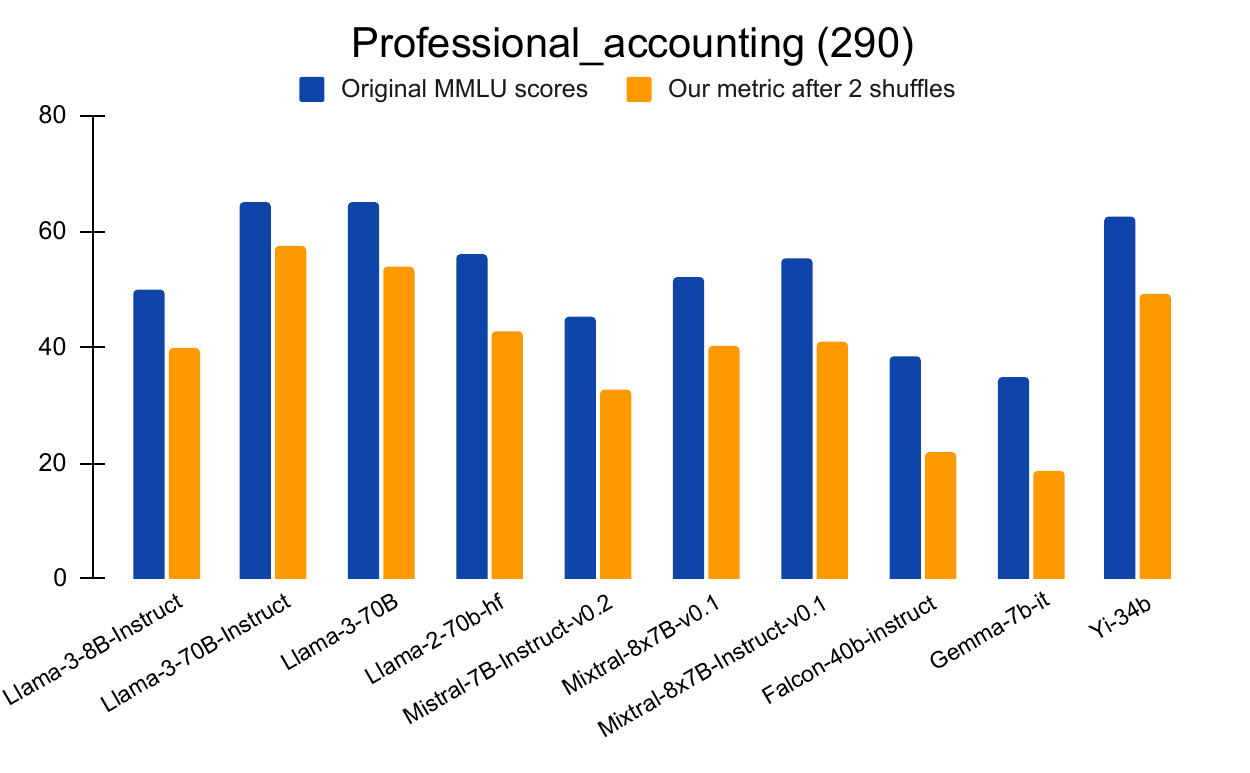}
  \end{subfigure}
  \caption{Here we show accuracy scores on random categories of MMLU with our proposed shuffling. The number along with each category name signifies the number of questions for that category in MMLU.}
  \label{fig:appenedix_plots}
\end{figure*}

\subsection {Computation Resources}

For all experiments for this work, we utilized 8 V100 32GB GPUs. These GPUs were assembled in a cluster of 8 GPUs in a node. The cumulative computing time required to evaluate all the language models and complete the experiments amounted to approximately 2000 GPU hours.

\end{document}